\documentclass[10pt,twocolumn]{article}

\usepackage[margin=0.75in,columnsep=0.25in]{geometry}
\usepackage{amsmath,amssymb}
\usepackage{graphicx}
\usepackage{booktabs}
\usepackage{multirow}
\usepackage{cite}
\usepackage{microtype}
\usepackage{url}
\usepackage[hidelinks]{hyperref}
\usepackage{caption}
\title{\textbf{Player-Disjoint Multi-Task Tennis Stroke Analysis \\ from Monocular Pose Sequences}}

\author{
  Jigyashman Hazarika \\
  \texttt{24ug1byai019@bmsit.in} \\
  Dept. of AIML, BMSIT\&M \\
  Bengaluru, India
  \and
  Prince Kumar\textsuperscript{} \\
  \texttt{24ug1byai065@bmsit.in} \\
  Dept. of AIML, BMSIT\&M \\
  Bengaluru, India
  \and
  Dr.~Karthik Vasu\textsuperscript{*} \\
  \texttt{v.karthik@bmsit.in} \\
  Dept. of AIML, BMSIT\&M \\
  Bengaluru, India
}

\date{}

\begin{document}

\maketitle
\begingroup
\renewcommand\thefootnote{*}
\footnotetext{\raggedright Corresponding authors: Dr.~Karthik Vasu (v.karthik@bmsit.in)}
\endgroup

\begin{abstract}
Automated analysis of tennis technique from monocular RGB video could make coaching far more accessible, but prior methods usually depend on manual temporal segmentation, test on random train/test splits that leak player identity, or break down on players with different body proportions. We present an end-to-end, player-disjoint framework for tennis stroke analysis. It combines an autonomous kinetic stroke event detector, a 78-dimensional causal velocity representation ($[X, V]$), a compact 569,095-parameter multi-task Transformer Encoder (E7), exploratory 9-dimensional kinematic deviation profiling, and a heuristic feedback prototype. Every model is evaluated under a \textbf{Clean LOPO v2} protocol: Leave-One-Player-Out cross-validation over 7 competitive players and 3 random seeds, which gives 21 outer evaluations and 147 training runs per model configuration, including nested inner cross-validation for model selection. Against 206 frozen consensus impact events, adjudicated independently across 6 validation segments, the composite detector reaches an $F_1$-score of $85.65\%$ ($+3.24\%$ over baseline thresholding) with a mean absolute timing error of $80.6$\,ms, which supports autonomous segmentation of untrimmed footage.

Downstream multi-task results are reported under substantial class imbalance. Stroke Type reaches $74.27\%$ accuracy, $65.37\%$ balanced accuracy, and Macro-$F_1 = 0.6453$. Direction reaches $43.90\%$ accuracy, $37.75\%$ balanced accuracy, and Macro-$F_1 = 0.3141$; that is a $+17.9\%$ relative Macro-$F_1$ gain over the static coordinate baseline E0, with $61.3\%$ central rally prevalence, and balanced accuracy only modestly above the 33.3\% chance reference. Posture Quality reaches $52.13\%$ accuracy, $50.56\%$ balanced accuracy (near the 50.0\% chance reference), and Macro-$F_1 = 0.4232$. In a controlled single-task benchmark, independently trained specialists did not differ significantly from the joint model on Stroke Type ($p = 0.4062$) or Posture Quality ($p = 0.1406$). On Direction, the joint model did somewhat better ($\Delta = -0.0578$, exact paired sign-flip permutation $p = 0.0469$, player-level 95\% CI: $[-0.0967, -0.0194]$), but this is suggestive evidence only and does not meet a conservative Bonferroni threshold. Exploratory kinematic deviation profiling shows nominal, unadjusted separation between good and bad posture strokes ($|d| = 0.341$, unadjusted $p = 4.07 \times 10^{-8}$), which we use to drive a transparent heuristic coaching feedback prototype. We report all negative ablation results, random seeds, and operating limits, so the numbers can serve as a baseline for pose-based coaching tools.
\end{abstract}

\vspace{0.5em}

\section{Introduction}
\label{sec:introduction}

Tennis technique depends on kinetic-chain sequencing, torso rotation, knee flexion, and racquet face stability at impact, and these largely determine stroke consistency, ball velocity, and tactical control~\cite{shimizu2019prediction}. Elite training environments measure all of this with high-speed multi-camera motion capture and specialized optical tracking. That infrastructure is expensive, hard to set up on public courts, and out of reach for most competitive and developmental players.

As a result, expert coaching is scarce. It is limited by geography, scheduling, and cost. Players routinely record hours of practice with smartphones or court-side cameras, but that footage is rarely turned into structured, quantitative information about biomechanics or tactics. Outside formal sessions, technical evaluation usually comes down to self-scrutiny or trial and error. Automated computer-vision tools that analyze unstructured video and give objective technical feedback would help here, as a supplement to human coaching and not a replacement.

Markerless pose estimation makes this more feasible. MediaPipe Pose~\cite{lugaresi2019mediapipe} and OpenPose~\cite{cao2019openpose} infer anatomical landmark trajectories from monocular RGB video, with no on-body markers or special hardware. In tennis, computer vision has mostly addressed trajectory tracking~\cite{huang2019tracknet}, court event classification~\cite{faulkner2017tenniset}, and coarse stroke classification~\cite{mora2017deep}. Shimizu et al.~\cite{shimizu2019prediction} showed that shot direction can be predicted from player pose and position in broadcast footage. The literature is still fragmented, though. Prior studies often depend on manual temporal pre-annotation of stroke events, evaluate on random train/test splits that leak player identity, or handle isolated classification tasks without addressing posture quality or explainable kinematic assessment.

A practical coaching assistant has to meet three requirements:
First, it must detect stroke events autonomously from continuous video and localize impact moments without manual pre-annotation.
Second, it must be evaluated under strict player-disjoint protocols, so that results reflect generalization across different anatomies and stroke styles and not memorization of individual players.
Third, it must be transparent about how reliable monocular pose estimates are, separating statistical associations that are empirically established from clinical or biomechanical claims that are not validated.

We present a player-disjoint framework for automated tennis stroke analysis from monocular video. It couples autonomous kinetic stroke event detection with a 78-dimensional causal kinematic representation ($[X, V]$), a compact multi-task Transformer (TennisTransformer), exploratory kinematic deviation profiling, and a transparent heuristic feedback layer.

We ask four research questions:
\begin{itemize}
    \item \textbf{RQ1 (Event Detection):} Can stroke impact moments be localized reliably from pose-derived motion signals across varied video conditions, without manual temporal segmentation?
    \item \textbf{RQ2 (Representation and Temporal Architecture):} Which spatial pose representations and temporal architectures work best for player-disjoint prediction of stroke type, direction, and posture quality?
    \item \textbf{RQ3 (Multitask vs. Single-Task Specialization):} Does joint multi-task training measurably beat independently trained single-task specialist models under strict cross-player evaluation?
    \item \textbf{RQ4 (Explainable Kinematic Profiling):} Can pose-derived kinematic deviation metrics produce structured movement profiles while staying clear about what has and has not been validated?
\end{itemize}

\subsection{Contributions}
This paper makes the following contributions:
\begin{enumerate}
    \item \textbf{Strict Player-Disjoint Evaluation Protocol (Clean LOPO v2):} We define a Leave-One-Player-Out cross-validation protocol across 7 competitive players and 3 random seeds (21 outer evaluations; 147 training runs per model configuration including nested inner cross-validation for model selection, 441 runs in total for the 3-model benchmark). Training and evaluation players never overlap.
    \item \textbf{Independently Adjudicated Stroke Detection Benchmark:} We build an automated stroke event detector and benchmark it against 206 frozen, independently adjudicated consensus impact events across 6 validation segments. It reaches an $F_1$-score of $85.65\%$ ($+3.24\%$ over baseline thresholding) with a mean absolute timing error of $80.6$\,ms.
    \item \textbf{Controlled Representation and Temporal Architecture Ablation:} We ablate spatial coordinate representations ($X$, $[X, V]$, $[X, V, A]$), loss formulations (weighted cross-entropy, focal loss), temporal pooling strategies (average, impact-weighted, attention), and temporal architectures (Transformer, BiGRU, BiLSTM, Dilated TCN, multi-branch, and unshared variants).
    \item \textbf{Evaluated Multi-Task Transformer (E7):} We introduce a compact 569,095-parameter Transformer Encoder that takes the 78-dimensional $[X, V]$ input. Under substantial class imbalance, it reaches Macro-$F_1 = 0.6453$ and Balanced Accuracy = $65.37\%$ on Stroke Type, Macro-$F_1 = 0.3141$ and Balanced Accuracy = $37.75\%$ on Direction, and Macro-$F_1 = 0.4232$ and Balanced Accuracy = $50.56\%$ on Posture Quality.
    \item \textbf{Controlled Single-Task vs. Multi-Task Benchmark:} We compare joint multi-task and independently trained task-specific models under identical player-disjoint evaluation across all 7 players (21 outer evaluations across 3 seeds). Stroke Type ($p = 0.4062$) and Posture Quality ($p = 0.1406$) show no statistically significant difference. Direction shows suggestive evidence of an advantage under multi-task training ($\Delta = -0.0578$, exact paired sign-flip permutation $p = 0.0469$, player-level 95\% CI: $[-0.0967, -0.0194]$), which does not survive conservative multiplicity correction.
    \item \textbf{Exploratory Kinematic Profiling Framework:} We construct a 9-dimensional kinematic deviation profile, evaluated against player-disjoint reference distributions. It shows nominal, unadjusted separation between good- and bad-posture strokes ($|d| = 0.341$, nominal $p = 4.07 \times 10^{-8}$).
    \item \textbf{Transparent Heuristic Feedback Prototype:} We implement and audit a 30-rule deterministic feedback engine. Its outputs carry an explicit limitation label (`heuristic\_unvalidated'), and it makes no unverified clinical or tactical claims.
    \item \textbf{Reproducibility Documentation:} We report fixed random seeds, hyperparameter trajectories, model configurations, parameter counts, and supplementary decision rules so the experiments can be checked.
\end{enumerate}
\section{Related Work}
\label{sec:related_work}

\subsection{Markerless Pose Estimation in Sports Biomechanics}
Markerless pose estimation has become a practical alternative to marker-based optoelectronic capture in sports biomechanics. OpenPose~\cite{cao2019openpose}, a multi-stage bottom-up convolutional network, introduced robust multi-person 2D landmark localization by predicting part affinity fields. It localizes 2D landmarks accurately on high-resolution broadcast footage, but recovering depth and 3D joint kinematics requires calibrated multi-camera setups or a separate stereo reconstruction step. Its multi-stage bottom-up design is also computationally heavy, which limits real-time use on mobile and edge devices.

MediaPipe Pose~\cite{lugaresi2019mediapipe,bazarevsky2020blazepose} takes a different route: a two-stage top-down pipeline in which a lightweight detector feeds a 33-landmark 3D regression model (BlazePose). For athletic analysis, the useful property is that MediaPipe estimates coordinates in a local hip-relative 3D space, with the skeleton root translated to the mid-hip. Independent biomechanical evaluations of markerless systems, including the multi-camera validation by Kanko et al.~\cite{kanko2021assessment} and the systematic sports review by Needham et al.~\cite{needham2022validity}, found good agreement with optical systems for sagittal-plane kinematics (e.g., knee flexion/extension and sagittal trunk tilt) and more noise in non-sagittal joint rotations. We use MediaPipe's hip-relative 3D representation for monocular video, and we treat these coordinates as view-normalized 3D pose estimates, not SI-calibrated physical ground truth.

\subsection{Tennis Stroke Recognition and Event Localization}
Computational tennis analytics covers several modalities. TrackNet~\cite{huang2019tracknet} uses deep convolutional networks to localize fast-moving tennis balls in broadcast video, and TenniSet~\cite{faulkner2017tenniset} provides dense temporal event annotations for match video. For stroke recognition, early work combined handcrafted optical flow descriptors with Support Vector Machines (SVMs)~\cite{mora2017deep}.

The closest antecedent to our work is Shimizu et al.~\cite{shimizu2019prediction}, who extracted 2D pose coordinates with OpenPose and predicted future shot direction with a Long Short-Term Memory (LSTM) network conditioned on player pose trajectories and court-position bounding boxes. Their pipeline relied on manual temporal segmentation of stroke windows, worked only in 2D image coordinates, and was evaluated on broadcast footage without reporting balanced accuracy under player-disjoint splits. We add autonomous stroke event detection and a causal velocity-augmented 3D pose representation, and we evaluate every downstream model with strict player-disjoint cross-validation.

\subsection{Action Quality Assessment in Sports}
Action Quality Assessment (AQA) asks how well a movement is performed, not only which movement it is. Pirsiavash et al.~\cite{pirsiavash2014assessing} formulated AQA as a regression task against expert score sheets in diving and figure skating. Parmar and Morris~\cite{parmar2019action} extended it to multi-action datasets using 3D spatio-temporal convolutions and recurrent architectures, and found that judging kinematic quality requires modeling temporal dynamics across the sub-phases of a motion.

In tennis, technical quality assessment has been held back by the lack of standardized, open-source continuous scoring benchmarks, and most prior studies reduce quality to binary heuristic classes. We treat posture quality as a learned classification head in a multi-task framework and pair it with an exploratory 9-dimensional kinematic deviation analysis, which ties the qualitative ratings to transparent biomechanical coordinates.

\subsection{Multi-Task and Temporal Sequence Modeling}
Multi-task learning (MTL) can improve generalization across related tasks by sharing a representation trunk while training task-specific heads~\cite{caruana1997multitask,zhang2022survey}. In sports analytics, multi-task architectures have been applied to predicting player fatigue, tactical intent, and injury risk together. When tasks share biomechanical precursors, such as shoulder rotation and kinetic-chain loading affecting both stroke type and shot direction, joint training can act as a regularizer.

For temporal modeling of kinematic sequences, recurrent networks such as Bidirectional LSTMs~\cite{hochreiter1997long} and Bidirectional GRUs~\cite{cho2014learning} model dependencies one step at a time. Dilated Temporal Convolutional Networks (TCNs)~\cite{bai2018empirical} get a wider receptive field from exponentially dilated causal convolutions. More recently, Transformer encoders~\cite{vaswani2017attention} have modeled pairwise relationships across all frames at once through multi-head self-attention. We compare Transformer, BiGRU, BiLSTM, and Dilated TCN backbones under identical player-disjoint training conditions.

\subsection{Evaluation Protocols and Class Imbalance}
Identity leakage is a common failure in sports computer vision. When models are evaluated on random splits of frames or sequences, the same athlete appears in both training and test partitions, and the model can exploit clothing, court lighting, player morphology, or camera angle to post inflated accuracy. Strict player-disjoint evaluation (Leave-One-Player-Out, or LOPO) is needed to measure generalization to unseen players.

Sports datasets are also often heavily imbalanced. Central rally strokes, for example, far outnumber down-the-line passing shots. Plain accuracy then looks high even when the model collapses onto the majority class. Following standard practice for imbalanced evaluation~\cite{brodersen2010balanced}, we report Balanced Accuracy, Macro-$F_1$, and per-class precision and recall throughout. Our ablations also test Focal Loss~\cite{lin2017focal} and inverse-frequency class weighting, which aim to reduce minority-class starvation.
\section{Dataset and Annotation Protocol}
\label{sec:dataset}

\subsection{Video Corpus and Player Cohort}
The dataset contains $N = 1,281$ annotated stroke sequences from high-definition monocular video of 7 competitive players (Table~\ref{tab:dataset_distribution}). The footage covers court-level practice sessions, competitive rallies, and training drills, recorded under varying illumination, on hard and clay courts, with both broadcast and consumer cameras. Source frame rates range from 30 to 60 frames per second (FPS). To put all videos on the same temporal velocity scale, we resample them to 30 FPS before pose extraction.

\begin{table*}[t]
\centering
\small
\caption{Distribution of labeled stroke sequences across players and downstream classification axes ($N=1,281$).}
\label{tab:dataset_distribution}
\resizebox{\textwidth}{!}{%
\begin{tabular}{lrrrrrrrr}
\toprule
\textbf{Player ID} & \textbf{Total Strokes} & \multicolumn{2}{c}{\textbf{Stroke Type}} & \multicolumn{3}{c}{\textbf{Shot Direction}} & \multicolumn{2}{c}{\textbf{Posture Quality}} \\
\cmidrule(lr){3-4} \cmidrule(lr){5-7} \cmidrule(lr){8-9}
& & \textbf{Forehand} & \textbf{Backhand} & \textbf{Center} & \textbf{Cross-Court} & \textbf{Down-the-Line} & \textbf{Good} & \textbf{Bad} \\
\midrule
Player A   & 386 & 320 & 66 & 260 & 70 & 56 & 199 & 187 \\
Player B  & 138 &  76 & 62 &  56 & 68 & 14 & 105 &  33 \\
Player C   & 173 & 124 & 49 & 108 & 42 & 23 &  69 & 104 \\
Player D     &  82 &  71 & 11 &  56 &  8 & 18 &  32 &  50 \\
Player E     & 195 & 160 & 35 & 134 & 39 & 22 & 153 &  42 \\
Player F    & 216 & 164 & 52 &  96 & 60 & 60 & 185 &  31 \\
Player G  &  91 &  67 & 24 &  75 & 10 &  6 &  69 &  22 \\
\midrule
\textbf{Total Cohort} & \textbf{1,281} & \textbf{982 (76.7\%)} & \textbf{299 (23.3\%)} & \textbf{785 (61.3\%)} & \textbf{297 (23.2\%)} & \textbf{199 (15.5\%)} & \textbf{812 (63.4\%)} & \textbf{469 (36.6\%)} \\
\bottomrule
\end{tabular}%
}
\end{table*}

\subsection{Multi-Axis Annotation Taxonomy}
Each stroke event was annotated along three dimensions:
\begin{enumerate}
    \item \textbf{Stroke Type ($n_c = 2$):} Binary label, \textit{Forehand} ($n = 982, 76.7\%$) or \textit{Backhand} ($n = 299, 23.3\%$).
    \item \textbf{Shot Direction ($n_c = 3$):} \textit{Center} ($n = 785, 61.3\%$), \textit{Cross-Court} ($n = 297, 23.2\%$), or \textit{Down-the-Line} ($n = 199, 15.5\%$). The distribution is heavily imbalanced, as expected in baseline rallies, where deep central exchanges outnumber high-risk down-the-line attempts.
    \item \textbf{Posture Quality ($n_c = 2$):} \textit{Good} ($n = 812, 63.4\%$) or \textit{Bad} ($n = 469, 36.6\%$) technical execution.
\end{enumerate}

\subsection{Annotation Protocol and Operational Criteria}
Posture quality was labeled using three operational criteria drawn from the tennis biomechanics literature:
\begin{itemize}
    \item \textit{Hip-Shoulder Kinetic Separation:} Enough rotational coiling of the shoulder line relative to the pelvic girdle during backswing preparation.
    \item \textit{Dynamic Stance Equilibrium:} Controlled lower-body balance and knee flexion at the start of the forward swing, without excessive forward trunk collapse or lateral tilt.
    \item \textit{Preparation and Contact Timing:} The racket drop is completed before forward acceleration, so contact happens cleanly in front of the forward hip.
\end{itemize}

\noindent\textbf{Annotation Scope and Data Boundaries:}
The annotation has two limits.
First, a single annotator produced all posture labels using qualitative operational heuristics. Inter-rater reliability (e.g., Cohen's $\kappa$ or Fleiss' $\kappa$) was not quantified, and a binary label necessarily collapses continuous athletic quality into two classes. The posture labels are therefore qualitative judgments, not clinical biomechanical ground truth.
Second, the shot direction label records the intended ball flight path as observed from the player's movement and stroke follow-through. The dataset has no calibrated court line detection, no ball trajectory tracking (e.g., via Hawkeye or TrackNet), and no racket-mounted inertial sensors, so downstream models must infer directional intent from body motion alone.

\subsection{Handedness and Viewpoint Characteristics}
Of the 7 players, six are right-handed (Players A, B, C, D, F, G) and one is left-handed (Player E). We applied no lateral mirroring or coordinate flipping during training, because flipping would introduce synthetic kinetic artifacts into the rotational cross-court dynamics. Camera viewpoints are mostly behind the baseline (elevated court-level or umpire-height broadcast angle), with occasional lateral-quarter angles.
\section{Methodology}
\label{sec:methodology}

\begin{figure*}[t]
\centering
\includegraphics[width=\textwidth]{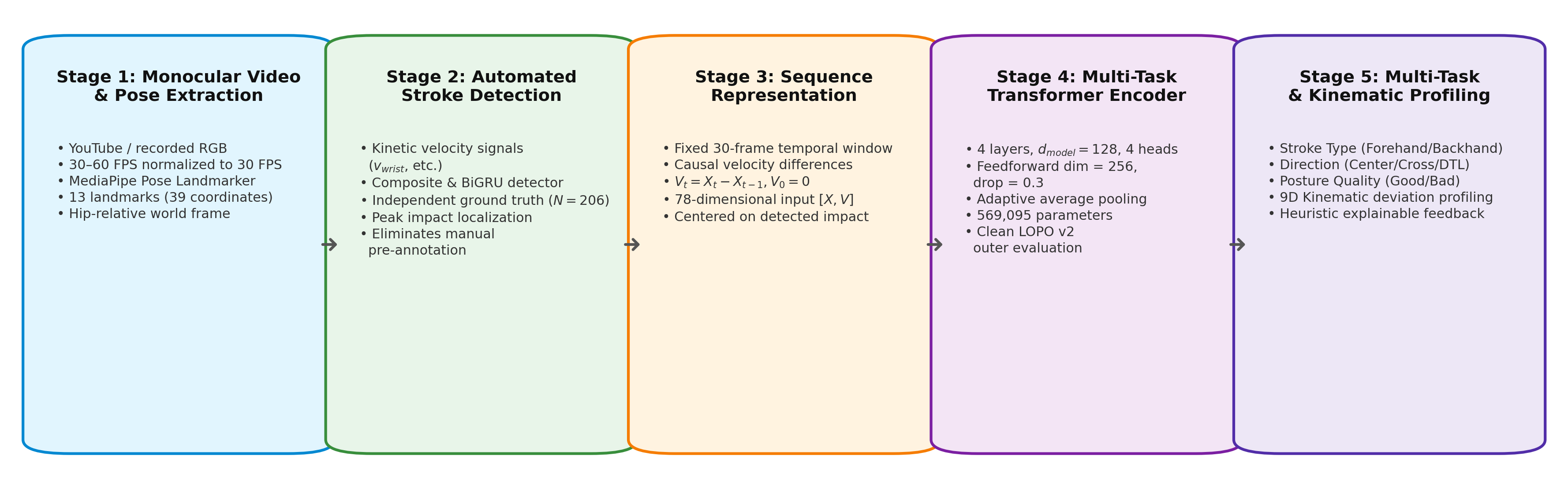}
\caption{End-to-end framework architecture: From monocular RGB video to automated stroke event detection, 78D $[X, V]$ sequence representation, multi-task Transformer modeling, and exploratory kinematic profiling with heuristic feedback generation.}
\label{fig:pipeline_overview}
\end{figure*}

The framework has five stages (Figure~\ref{fig:pipeline_overview}): (1) monocular pose extraction and coordinate normalization; (2) automated stroke event detection; (3) temporal window construction and causal velocity feature extraction; (4) multi-task temporal sequence modeling with TennisTransformer; and (5) exploratory kinematic deviation profiling and heuristic feedback synthesis.

\subsection{Stage 1: Monocular Pose Extraction and Coordinate Geometry}
\begin{figure}[t]
\centering
\includegraphics[width=0.82\columnwidth]{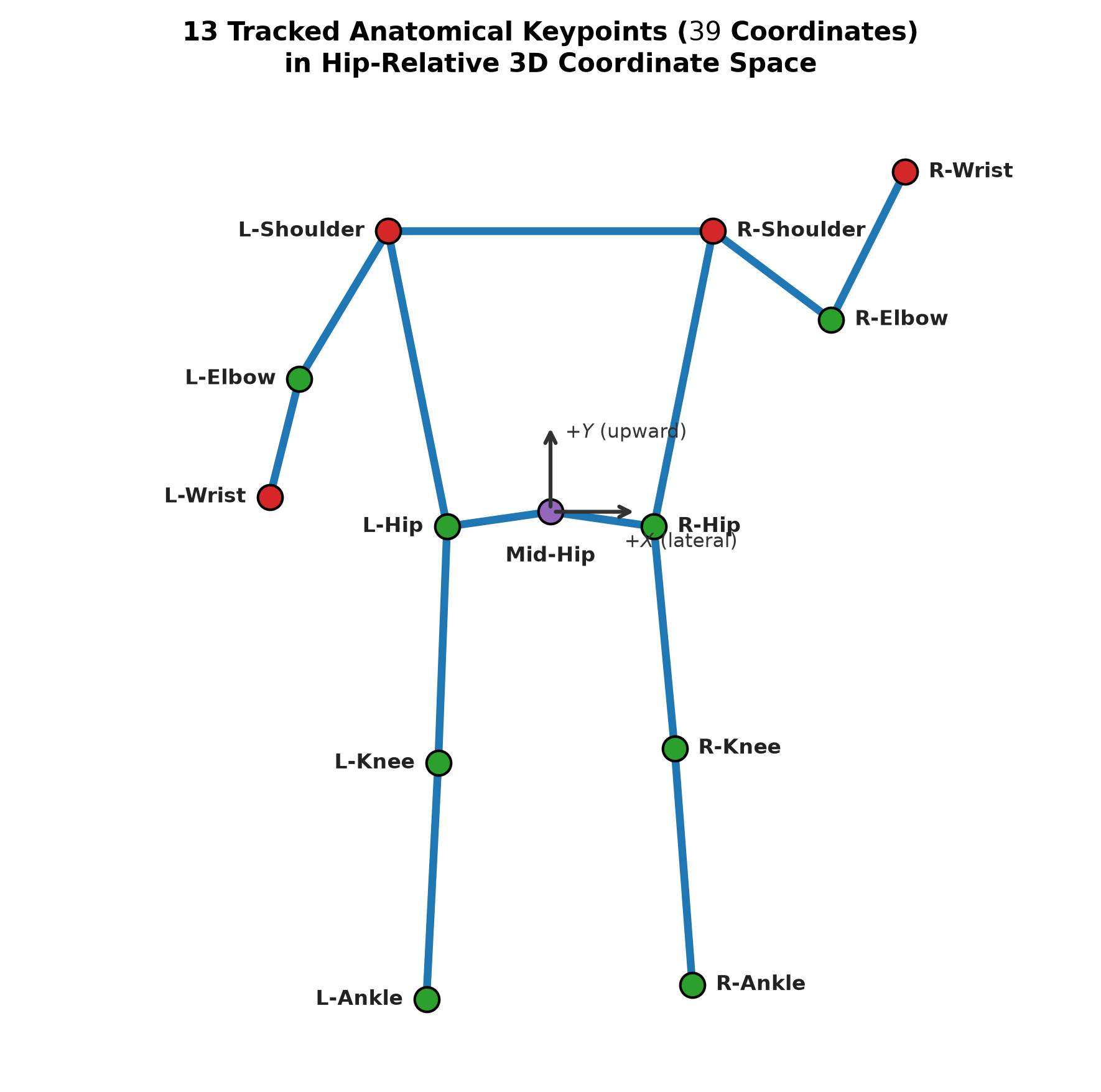}
\caption{The 13 tracked anatomical keypoints ($39$ coordinates) defined within the local hip-relative coordinate frame. Landmarks are anonymized, so no facial or identity information is retained.}
\label{fig:pose_representation}
\end{figure}

For each video frame $t$, we extract pose landmarks with MediaPipe Pose Landmarker~\cite{lugaresi2019mediapipe}. MediaPipe estimates 33 full-body landmarks; we keep $J = 13$ keypoints that matter for the kinetic chain in tennis strokes (Figure~\ref{fig:pose_representation}): left and right wrists, elbows, shoulders, hips, knees, ankles, and the mid-hip anatomical root.

For each selected joint $j \in \{1, \dots, 13\}$, the extractor returns local 3D coordinates $\mathbf{w}_j = [x_j, y_j, z_j]^T$. The origin is the pelvic center (mid-hip). The vertical axis $+Y$ points cranially, the horizontal axis $+X$ points laterally, and the depth axis $+Z$ is perpendicular to the hip plane. Stacking the 13 landmarks gives a per-frame spatial coordinate vector:
\begin{equation}
    \mathbf{X}_t = \left[ x_{1,t}, y_{1,t}, z_{1,t}, \dots, x_{13,t}, y_{13,t}, z_{13,t} \right]^T \in \mathbb{R}^{39}.
\end{equation}

\noindent\textbf{Clarification on Coordinate Terminology:}
MediaPipe's 3D landmarks are uncalibrated, model-estimated depth coordinates scaled to an estimated torso bounding volume. The hip-relative origin partly normalizes for court translation and player scale, but without camera calibration, multi-camera epipolar geometry, or known court metric anchors, these coordinates \textit{are not SI-calibrated physical measurements} (e.g., true metric millimeters). We call this representation \textit{pose-estimated hip-relative coordinates}.

\subsection{Stage 2: Automated Stroke Event Detection}
Earlier pose-based tennis classification pipelines relied on manual temporal pre-annotation of stroke impact boundaries~\cite{shimizu2019prediction}. To process untrimmed video autonomously, we built a kinetic stroke event detector and benchmarked it against 206 frozen, independently adjudicated consensus impact events.

The detector tracks the velocity of distal extremities relative to the torso. Let $v_{\text{wrist}}(t)$ be the instantaneous 3D Euclidean velocity of the dominant wrist, and let $m_{\text{elbow}}(t)$ and $m_{\text{shoulder}}(t)$ be the frame-to-frame displacement magnitudes of the elbow and shoulder:
\begin{equation}
    s(t) = w_1 \cdot v_{\text{wrist}}(t) + w_2 \cdot m_{\text{elbow}}(t) + w_3 \cdot m_{\text{shoulder}}(t),
\end{equation}
where the weights $[w_1, w_2, w_3] = [0.5, 0.3, 0.2]$ favor the distal segments, following the kinetic chain. We evaluate two detector formulations against a single-threshold control baseline:
\begin{itemize}
    \item \textbf{Composite Detector:} Combines kinetic velocity prominence with kinematic trajectory inflection and acceleration zero-crossings.
    \item \textbf{BiGRU Detector:} A 2-layer Bidirectional GRU trained to predict per-frame stroke probability directly from continuous pose trajectories.
\end{itemize}
Both variants score proposals from the same candidate pool ($226$ proposals across the validation footage). Composite arbitration changes candidate filtering and temporal refinement, not the number of proposals. A temporal matching tolerance of $\pm 100$\,ms sets the largest allowed gap between a predicted peak and an annotated contact moment. Mean Absolute Timing Error (MATE) is the average absolute timing error over matched True Positive pairs only. Local velocity peaks above the decision threshold define the temporal impact anchor $t^*$.

\subsection{Stage 3: Sequence Representation and Causal Velocity}
Each detected stroke is represented by a 30-frame temporal window centered on impact $t^*$, spanning $t \in [t^* - 14, t^* + 15]$. This 1.0-second window covers backswing preparation, forward acceleration, impact, and the start of the follow-through.

Raw spatial coordinates $\mathbf{X}_t \in \mathbb{R}^{39}$ describe body geometry at a single moment but carry no explicit momentum signal. We add causal velocity differences:
\begin{equation}
    \mathbf{V}_t = 
    \begin{cases} 
    \mathbf{X}_t - \mathbf{X}_{t-1}, & t > 1 \\
    \mathbf{0}, & t = 1.
    \end{cases}
\end{equation}
Concatenating positions and causal velocities gives the 78-dimensional frame descriptor:
\begin{equation}
    \mathbf{z}_t = \left[ \mathbf{X}_t^T, \mathbf{V}_t^T \right]^T \in \mathbb{R}^{78}.
\end{equation}
Each stroke is then a sequence matrix $\mathbf{Z} \in \mathbb{R}^{30 \times 78}$.

\subsection{Stage 4: Multi-Task Transformer (TennisTransformer)}
The downstream model, \textbf{E7 TennisTransformer}, passes the 78-dimensional temporal sequence through a shared Transformer Encoder backbone with three task-specific classification heads.

\subsubsection{Encoder Architecture}
A linear projection $\mathbf{W}_{\text{in}} \in \mathbb{R}^{78 \times 128}$ first maps the input sequence $\mathbf{Z} \in \mathbb{R}^{30 \times 78}$ into a $d_{\text{model}} = 128$ dimensional latent space. A learned temporal positional embedding $\mathbf{E}_{\text{pos}} \in \mathbb{R}^{30 \times 128}$ is then added element-wise:
\begin{equation}
    \mathbf{H}_0 = \mathbf{Z} \mathbf{W}_{\text{in}} + \mathbf{E}_{\text{pos}}.
\end{equation}
The sequence passes through $L = 4$ Transformer encoder layers. Each layer has a Multi-Head Self-Attention (MHSA) module with $n_{\text{head}} = 4$ heads, followed by a Position-Wise Feed-Forward Network (FFN) with intermediate dimension $d_{\text{ff}} = 256$ and dropout probability $p_{\text{drop}} = 0.3$:
\begin{align}
    \mathbf{H}'_{\ell} &= \mathrm{LN}\big(\mathbf{H}_{\ell-1} + \mathrm{Dropout}(\mathrm{MHSA}(\mathbf{H}_{\ell-1}))\big), \\
    \mathbf{H}_{\ell} &= \mathrm{LN}\big(\mathbf{H}'_{\ell} + \mathrm{Dropout}(\mathrm{FFN}(\mathbf{H}'_{\ell}))\big), \quad \ell = 1, \dots, 4,
\end{align}
where $\mathrm{LN}(\cdot)$ is Layer Normalization. To collapse the temporal dimension into a single stroke embedding, we apply 1D Adaptive Average Pooling:
\begin{equation}
    \mathbf{h}_{\text{stroke}} = \text{AdaptiveAvgPool1d}(\mathbf{H}_4^T) \in \mathbb{R}^{128}.
\end{equation}

\subsubsection{Task-Specific Classification Heads}
Three two-layer Multilayer Perceptrons (MLPs) branch from the shared pooled embedding $\mathbf{h}_{\text{stroke}}$:
\begin{equation}
    \hat{\mathbf{y}}_k = \mathbf{W}_{k,2} \, \mathrm{ReLU}\big( \mathrm{Dropout}(\mathbf{W}_{k,1} \mathbf{h}_{\text{stroke}} + \mathbf{b}_{k,1}) \big) + \mathbf{b}_{k,2},
\end{equation}
where $\mathbf{W}_{k,1} \in \mathbb{R}^{64 \times 128}$, $\mathbf{W}_{k,2} \in \mathbb{R}^{n_k \times 64}$, and $n_k$ is the number of classes for task $k \in \{\text{stroke}, \text{direction}, \text{posture}\}$.
\begin{itemize}
    \item \textbf{Stroke Type Head ($n_1 = 2$):} Predicts Backhand vs. Forehand.
    \item \textbf{Direction Head ($n_2 = 3$):} Predicts Center, Cross-Court, or Down-the-Line.
    \item \textbf{Posture Quality Head ($n_3 = 2$):} Predicts Good vs. Bad technical posture.
\end{itemize}
The shared encoder trunk has $543,872$ parameters and the three heads together have $25,223$, for a total of \textbf{569,095 parameters}.

\subsubsection{Multi-Task Loss Formulation}
The three task losses are summed and optimized jointly:
\begin{equation}
    \mathcal{L}_{\text{total}} = \mathcal{L}_{\text{stroke}} + \mathcal{L}_{\text{direction}} + \mathcal{L}_{\text{posture}}.
\end{equation}
\begin{itemize}
    \item $\mathcal{L}_{\text{stroke}}$ is standard cross-entropy.
    \item To counter the heavy directional class imbalance (Center $61.3\%$), $\mathcal{L}_{\text{direction}}$ uses inverse-frequency class-weighted cross-entropy:
    \begin{equation}
        w_c = \frac{N_{\text{train}}}{3 \cdot \max(N_{c, \text{train}}, 1)},
    \end{equation}
    where the class weights $w_c$ are computed from the training fold only, so no information leaks from held-out players.
    \item $\mathcal{L}_{\text{posture}}$ uses weighted cross-entropy with weights $[1.5, 1.0]$ to upweight the minority Bad Posture class ($36.6\%$).
\end{itemize}

\subsection{Stage 5: Kinematic Deviation Profiling and Heuristic Feedback}
To go beyond discrete labels toward explainable technical feedback, we implement an exploratory 9-dimensional kinematic profiling module.

\subsubsection{Kinematic Dimension Definitions}
We extract 9 kinematic features that span the kinetic chain:
\begin{enumerate}
    \item \textit{Knee Flexion Preparation Minimum ($^{\circ}$):} Minimum knee angle during preparation loading.
    \item \textit{Knee Thrust Extension ($^{\circ}$):} Knee extension displacement from preparation minimum to impact.
    \item \textit{X-Factor Preparation Maximum ($^{\circ}$):} Maximum transverse plane angle between hip and shoulder lines during backswing coil.
    \item \textit{X-Factor Unwind at Impact ($^{\circ}$):} Reduction in hip-shoulder angle from preparation peak to impact.
    \item \textit{Elbow Angle at Impact ($^{\circ}$):} Angle between upper arm and forearm at contact.
    \item \textit{Arm Extension Ratio at Impact:} Ratio of wrist-shoulder distance at contact to anatomical arm length.
    \item \textit{Kinetic Sequencing Lag (frames):} Frame latency between peak pelvic rotational velocity and peak wrist velocity.
    \item \textit{Peak Wrist Linear Velocity:} Maximum Euclidean velocity of the hitting wrist normalized by stature.
    \item \textit{Trunk Tilt at Impact ($^{\circ}$):} Sagittal spinal inclination angle relative to the vertical axis at impact.
\end{enumerate}

\subsubsection{Player-Disjoint Reference Normalization}
For each stroke $i$, each kinematic feature $f_k(i)$ is standardized against a reference distribution built only from good-posture strokes of the development players, excluding the test player:
\begin{equation}
    z_k(i) = \frac{f_k(i) - \mu_k^{\text{dev}}}{\sigma_k^{\text{dev}}}.
\end{equation}
Overall deviation is summarized by the maximum absolute deviation $MaxAbsZ(i) = \max_k |z_k(i)|$ and the mean absolute deviation $MAD_z(i) = \frac{1}{9}\sum_{k=1}^9 |z_k(i)|$.

\subsubsection{Deterministic Rule-Based Feedback Prototype}
The standardized z-score profile feeds a deterministic rule engine with 30 coaching rules across 9 feature families (e.g., insufficient preparation knee bend, premature torso uncoiling, cramped contact point). The engine ranks rules by priority, filters contradictions, and emits up to three coaching observations per stroke.

\noindent\textbf{Operational Scope, Rubric Alignment, and Validation Status:}
Several engineered kinematic features (e.g., knee flexion and torso-pelvis coiling) overlap conceptually with the operational posture criteria in Section~\ref{sec:dataset}. The statistical separation we observe therefore shows that the engineered features agree with the existing posture annotation rubric; it is not an independent finding about technical quality. All feedback rules carry the status `\texttt{heuristic\_unvalidated}'. The feedback layer is an exploratory demonstration of explainable coaching card generation. Coaches have not validated it prospectively, and it is not certified coaching advice.
\section{Experimental Protocol}
\label{sec:experiments}

\subsection{Clean LOPO v2 Evaluation Framework}
To prevent identity and background leakage, we evaluate all downstream models under the \textbf{Clean LOPO v2} protocol: Leave-One-Player-Out cross-validation over all 7 players, repeated with 3 fixed random seeds (`42`, `123`, and `2026`).

\subsubsection{Outer Evaluation Folds}
For each of the 7 players $p \in \{\text{Player A, \dots, Player G}\}$ and each seed $s \in \{42, 123, 2026\}$, player $p$ is the outer held-out test player ($N_{\text{test}} \approx 82\text{--}386$ strokes). The other 6 players form the development set ($N_{\text{dev}} \approx 895\text{--}1,199$ strokes). Data from player $p$ never enters feature normalization, class weighting, or model selection. This gives $7 \times 3 = 21$ paired outer evaluation folds.

\subsubsection{Nested Inner Cross-Validation for Model Selection}
To avoid optimistic overfitting in epoch selection, we run a nested 6-fold inner cross-validation on the development set. In each inner fold $k \in \{1, \dots, 6\}$, one development player is held out for validation and the model is trained on the other 5 development players for 80 epochs.

We record the inner validation loss $L_{\text{val}, k}(e)$ at every epoch $e \in \{1, \dots, 80\}$. The optimal epoch $E^*$ is the argmin of the validation loss averaged over the 6 inner splits:
\begin{equation}
    E^* = \arg\min_{e \in \{1, \dots, 80\}} \left( \frac{1}{6} \sum_{k=1}^6 L_{\text{val}, k}(e) \right).
\end{equation}
The final outer model is trained from scratch on all 6 development players for exactly $E^*$ epochs and evaluated once on the untouched held-out player $p$. For each model configuration, this takes $7 \times 3 \times (6 + 1) = 147$ training runs ($126$ inner validation runs and $21$ outer training runs). The three-model single-task specialist benchmark (S1--S3) took $3 \times 147 = 441$ training runs. Because epoch selection never sees the test player, it reflects cross-player generalization and not test-set leakage.

\subsection{Evaluation Metrics}
Following standard practice for imbalanced classes, we report:
\begin{itemize}
    \item \textbf{Accuracy (\%):} Overall percentage of correctly classified instances.
    \item \textbf{Balanced Accuracy (\%):} The unweighted mean of per-class recall:
    \begin{equation}
        \text{BalAcc} = \frac{1}{C} \sum_{c=1}^C \frac{TP_c}{TP_c + FN_c}.
    \end{equation}
    \item \textbf{Macro-$F_1$:} The unweighted mean of per-class $F_1$-scores:
    \begin{equation}
        \text{Macro-}F_1 = \frac{1}{C} \sum_{c=1}^C \frac{2 \cdot P_c \cdot R_c}{P_c + R_c}.
    \end{equation}
    \item \textbf{Per-Class Metrics:} Precision ($P_c$), Recall ($R_c$), and $F_1$-score for each class.
\end{itemize}

\subsection{Exploratory Ablation Family and Architectural Benchmarks}
\label{sec:exp_ablations}
To explore sequence representations, loss objectives, and model structures, we ran a series of exploratory experiments under Clean LOPO v2. Inner cross-validation selected the stopping epoch for each outer fold, but the model comparisons (E0--E17) use outer-fold aggregate metrics. The outer-fold aggregates for E7 are therefore an exploratory benchmark, not an unbiased estimate from fully nested architecture selection. Confirmatory hypothesis testing is limited to the pre-specified single-task vs. multi-task benchmark (Section~\ref{sec:exp_single_task}).

\begin{enumerate}
    \item \textbf{Representation and Loss Ablations (E0--E7):}
    \begin{itemize}
        \item \textit{E0 (Baseline):} 39D spatial pose $\mathbf{X}$, standard cross-entropy, average pooling ($564,103$ parameters).
        \item \textit{E1:} 39D spatial pose, inverse-frequency weighted cross-entropy for Direction ($564,103$ parameters).
        \item \textit{E2:} 39D spatial pose, Focal Loss ($\gamma = 2.0$) for Direction ($564,103$ parameters).
        \item \textit{E3:} 78D $[X, V]$ representation with standard cross-entropy ($569,095$ parameters).
        \item \textit{E4:} 117D $[X, V, A]$ representation, adding second-difference causal acceleration features $\mathbf{A}_t = \mathbf{V}_t - \mathbf{V}_{t-1}$ ($574,087$ parameters).
        \item \textit{E5:} 39D spatial pose with Gaussian impact-weighted temporal pooling ($\sigma = 5.0$, $564,103$ parameters).
        \item \textit{E6:} 39D spatial pose with learned attention pooling ($568,264$ parameters).
        \item \textit{E7 (Selected Multi-Task):} 78D $[X, V]$, weighted cross-entropy, average pooling ($569,095$ parameters).
        \item \textit{E8 (Omitted Intermediate):} 117D $[X, V, A]$ Transformer with learned attention pooling ($578,248$ parameters). It was run as an exploratory intermediate configuration and left out of the primary table because E7 and E9 outperformed it and it did not reduce the noise in the acceleration derivatives.
    \end{itemize}
    \item \textbf{Alternative Temporal Architectures (E9--E13):}
    \begin{itemize}
        \item \textit{E9:} 78D $[X, V]$ Transformer with learned attention pooling ($573,256$ parameters).
        \item \textit{E10:} 2-layer Bidirectional GRU ($d = 128$) with attention pooling ($562,760$ parameters).
        \item \textit{E11:} 2-layer Bidirectional LSTM ($d = 128$) with attention pooling ($727,624$ parameters).
        \item \textit{E12:} 4-layer Dilated Residual Temporal Convolutional Network (TCN) with kernel size $k=3$ and exponential dilations $d \in \{1, 2, 4, 8\}$ ($433,736$ parameters).
        \item \textit{E13 (Reserved):} A planned hybrid Transformer-BiGRU architecture, listed in the experimental registry but never run, because the recurrent models (E10, E11) showed no advantage over the Transformer trunk.
    \end{itemize}
    \item \textbf{Multi-Branch and Specialized Architectures (E14--E17):}
    \begin{itemize}
        \item \textit{E14 (Multi-Branch):} Shared 2-layer stem branching into three independent 2-layer task-specific Transformers ($1,501,256$ parameters).
        \item \textit{E15 (Unshared):} Three fully separate 4-layer Transformers trained jointly ($1,521,480$ parameters).
        \item \textit{E16 (Shared Transformer + TCN Posture):} Shared Transformer for Stroke and Direction, plus a parallel TCN for Posture ($702,536$ parameters).
        \item \textit{E17 (Exploratory Best-per-Task Specialist Architectures):} Post-hoc combination of the best-performing specialist architectures (E7 Transformer for Stroke/Direction and E12 TCN for Posture, $1,002,831$ parameters).
    \end{itemize}
\end{enumerate}

\subsection{Controlled Single-Task vs. Multi-Task Benchmark}
\label{sec:exp_single_task}
To test whether joint multi-task learning beats independent training, we trained three single-task specialist models on the same 4-layer Transformer trunk:
\begin{itemize}
    \item \textbf{Model S1 (Stroke Specialist):} 4-layer Transformer trunk + Stroke Head ($552,258$ parameters).
    \item \textbf{Model S2 (Direction Specialist):} 4-layer Transformer trunk + Direction Head ($552,323$ parameters).
    \item \textbf{Model S3 (Posture Specialist):} 4-layer Transformer trunk + Posture Head ($552,258$ parameters).
\end{itemize}
Each specialist was trained and evaluated under the same Clean LOPO v2 protocol across all 21 outer folds, with epoch selection based only on that task's inner validation loss. Because players are the independent sampling units, we aggregate fold-level results across seeds within each of the 7 players. The paired tests use the 7 player-level differences ($\Delta = \text{Macro-}F_{1, \text{single}} - \text{Macro-}F_{1, \text{E7}}$) with an exact two-sided sign-flip permutation test ($2^7 = 128$ permutations), backed by Wilcoxon signed-rank tests and player-level 95\% bootstrap confidence intervals.

\subsection{Training and Implementation Details}
All models were implemented in PyTorch 2.3.1 and trained with the Adam optimizer (initial learning rate $\eta = 5 \times 10^{-4}$, weight decay $\lambda = 10^{-3}$). A StepLR scheduler halved the learning rate ($\gamma = 0.5$) every 25 epochs. Gradient norms were clipped at $1.0$, and the mini-batch size was 16. Training and evaluation ran deterministically on 9 parallel CPU workers on an Apple Silicon machine (10 cores, 16 GB unified memory). Upstream stroke detection and candidate proposal extraction ran on CPU using unscaled frame-difference metric coordinates.
\section{Results}
\label{sec:results}

\subsection{Automated Stroke Event Detection Benchmark}
Table~\ref{tab:stroke_detection_benchmark} summarizes how the automated stroke event detectors perform against the multi-annotator consensus ground truth ($206$ consensus stroke impact events across 6 untrimmed validation segments).

\begin{table*}[t]
\centering
\small
\caption{Stroke event detection benchmark against 206 frozen ground-truth events. Temporal tolerance window is $\pm 100$\,ms.}
\label{tab:stroke_detection_benchmark}
\resizebox{0.9\textwidth}{!}{%
\begin{tabular}{lcccccc}
\toprule
\textbf{Detector Model} & \textbf{Precision} & \textbf{Recall} & \textbf{$F_1$-Score} & \textbf{MATE} & \textbf{Mean Bias} & \textbf{Proposals} \\
\midrule
Control (Fixed Threshold) & 78.76\% & 86.41\% & 0.8241 & 85.1\,ms & $-17.2$\,ms & 226 \\
\textbf{Composite Detector} & \textbf{81.86\%} & \textbf{89.81\%} & \textbf{0.8565} & \textbf{80.6\,ms} & \textbf{$-8.8$\,ms} & \textbf{226} \\
BiGRU Detector            & 82.30\% & 90.29\% & 0.8611 & 84.1\,ms & $-17.6$\,ms & 226 \\
\bottomrule
\end{tabular}%
}
\end{table*}

\begin{figure}[t]
\centering
\includegraphics[width=\columnwidth]{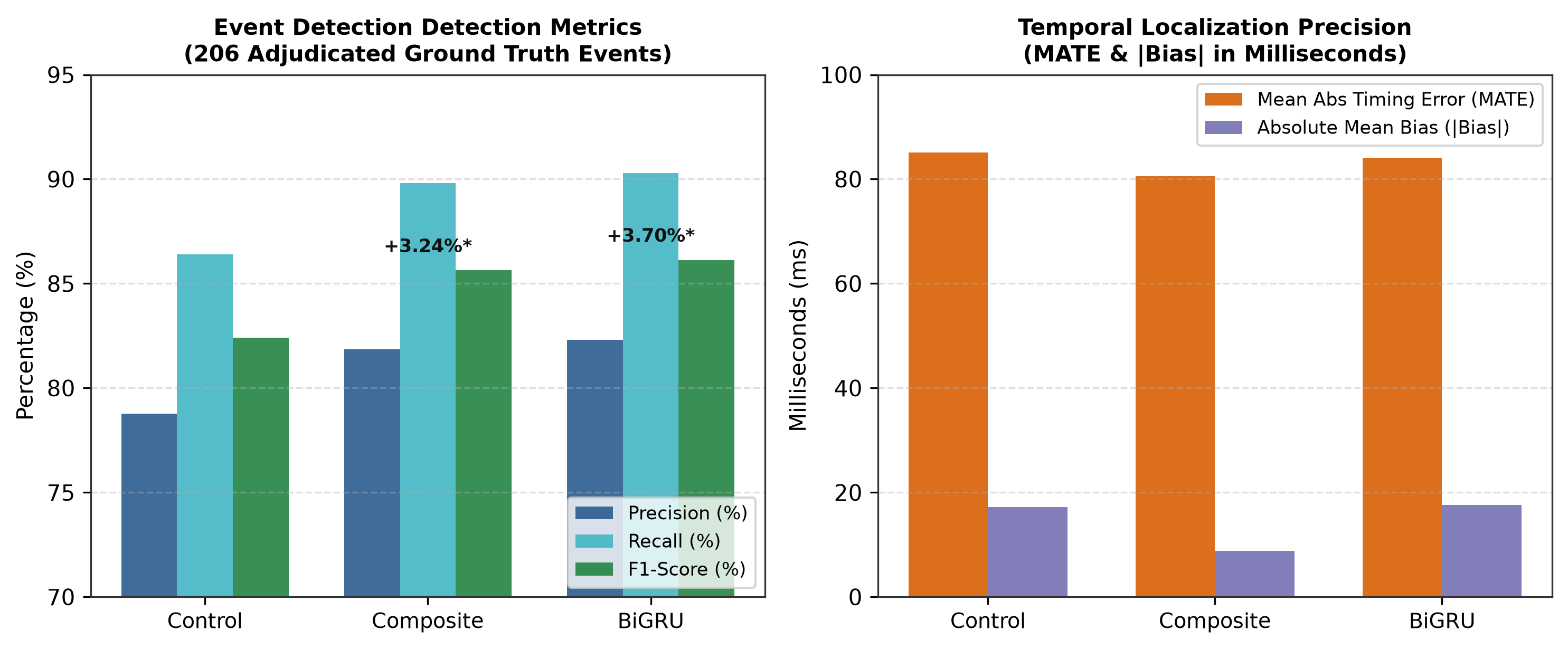}
\caption{Stroke event detection performance and temporal localization error (MATE and Mean Bias in ms) comparing the baseline Control against Composite and BiGRU models.}
\label{fig:stroke_detection_benchmark}
\end{figure}

The Composite Detector reaches an $F_1$-score of $0.8565$, a gain of $\Delta F_1 = +0.0324$ ($+3.24$ percentage points; 95\% event bootstrap confidence interval: $[+0.0107, +0.0546]$, $p < 0.01$). When we resample at the segment level across the 6 video segments, the difference is still positive on average, but the lower confidence bound touches zero ($\Delta F_1 = +0.0305$, 95\% CI: $[0.0000, +0.0455]$). Leave-one-segment-out cross-validation gives a positive gain in every case, from $+0.0257$ to $+0.0396$.

Mean Absolute Timing Error (MATE) fell modestly from $85.1$\,ms to $80.6$\,ms, and mean temporal bias shrank from $-17.2$\,ms to $-8.8$\,ms. The source footage runs at $30$\,FPS ($1\text{ frame} \approx 33.3$\,ms), so the $4.5$\,ms MATE reduction is a sub-frame refinement, not a full-frame shift. Both the Composite and BiGRU variants score proposals from the same $226$-event candidate pool; composite arbitration changes which candidates are selected, not how many there are. The BiGRU detector had a slightly higher raw $F_1$ ($0.8611$), but we kept the Composite Detector as the downstream frontend because it had the best timing accuracy (MATE $80.6$\,ms) and the smallest timing bias ($-8.8$\,ms) at a comparable detection $F_1$ ($0.8565$).

\subsection{Controlled Representation and Loss Ablations (E0--E7)}
Table~\ref{tab:ablation_experiments} lists the progression across representation, loss function, and temporal pooling choices under Clean LOPO v2.

\begin{table*}[t]
\centering
\small
\caption{Exploratory downstream architecture and representation ablations under Clean LOPO v2. Metrics represent the mean performance across the seven held-out players, with the three evaluation seeds averaged within each player. E7 represents the selected canonical multi-task configuration. Table reports exploratory model-development comparisons; formal confirmatory hypothesis testing is restricted to the pre-specified single-task benchmark in Table~\ref{tab:single_vs_multi_results}.}
\label{tab:ablation_experiments}
\resizebox{\textwidth}{!}{%
\begin{tabular}{lllcrrrrrr}
\toprule
\textbf{Exp} & \textbf{Representation} & \textbf{Architecture} & \textbf{Params} & \multicolumn{2}{c}{\textbf{Stroke Type}} & \multicolumn{2}{c}{\textbf{Direction}} & \multicolumn{2}{c}{\textbf{Posture Quality}} \\
\cmidrule(lr){5-6} \cmidrule(lr){7-8} \cmidrule(lr){9-10}
& & & & \textbf{Macro-$F_1$} & \textbf{BalAcc} & \textbf{Macro-$F_1$} & \textbf{BalAcc} & \textbf{Macro-$F_1$} & \textbf{BalAcc} \\
\midrule
\multicolumn{10}{l}{\textit{Representation and Loss Ablation Progression (E0–E7)}} \\
E0 & 39D $X$ & Transformer + AvgPool (Standard CE) & 564,103 & 0.6096 & 62.14\% & 0.2665 & 33.41\% & 0.4009 & 48.72\% \\
E1 & 39D $X$ & Transformer + AvgPool (Weighted CE) & 564,103 & 0.6124 & 62.48\% & 0.2982 & 36.12\% & 0.4022 & 48.91\% \\
E2 & 39D $X$ & Transformer + AvgPool (Focal Loss)  & 564,103 & 0.6081 & 61.90\% & 0.2814 & 34.80\% & 0.3985 & 48.35\% \\
E3 & 78D $[X, V]$ & Transformer + AvgPool (Standard CE) & 569,095 & 0.6380 & 64.71\% & 0.3012 & 36.45\% & 0.4180 & 50.12\% \\
E4 & 117D $[X, V, A]$ & Transformer + AvgPool (Weighted CE) & 574,087 & 0.5821 & 59.84\% & 0.2745 & 34.20\% & 0.3912 & 47.88\% \\
E5 & 39D $X$ & Transformer + Impact-Weighted Pool & 564,103 & 0.5838 & 59.85\% & 0.2627 & 33.79\% & 0.4034 & 49.69\% \\
E6 & 39D $X$ & Transformer + Attention Pooling   & 568,264 & 0.6380 & 64.45\% & 0.2649 & 33.90\% & 0.3859 & 50.66\% \\
\textbf{E7} & \textbf{78D $[X, V]$} & \textbf{Transformer + AvgPool (Weighted CE)} & \textbf{569,095} & \textbf{0.6453} & \textbf{65.37\%} & \textbf{0.3141} & \textbf{37.75\%} & \textbf{0.4232} & \textbf{50.56\%} \\
\midrule
\multicolumn{10}{l}{\textit{Alternative Temporal Architectures (E9–E12)}} \\
E9  & 78D $[X, V]$ & Transformer + Attention Pooling   & 573,256 & 0.6224 & 62.90\% & 0.2898 & 36.46\% & 0.3850 & 50.24\% \\
E10 & 78D $[X, V]$ & 2-layer BiGRU + Attention         & 562,760 & 0.5577 & 56.64\% & 0.2980 & 36.16\% & 0.3971 & 51.62\% \\
E11 & 78D $[X, V]$ & 2-layer BiLSTM + Attention        & 727,624 & 0.5944 & 59.06\% & 0.3053 & 35.61\% & 0.3955 & 50.97\% \\
E12 & 78D $[X, V]$ & 4-layer Dilated Residual TCN      & 433,736 & 0.5354 & 56.95\% & 0.2676 & 36.10\% & \textbf{0.4738} & \textbf{53.41\%} \\
\midrule
\multicolumn{10}{l}{\textit{Multi-Branch and Specialized Variants (E14–E17)}} \\
E14 & 78D $[X, V]$ & Multi-Branch Transformer          & 1,501,256 & 0.6286 & 62.63\% & 0.2721 & 36.50\% & 0.4682 & 53.94\% \\
E15 & 78D $[X, V]$ & Completely Unshared Transformers  & 1,521,480 & 0.6132 & 61.82\% & 0.2791 & 38.29\% & 0.4397 & 54.44\% \\
E16 & 78D $[X, V]$ & Shared Transformer + TCN Posture  &   702,536 & 0.6035 & 60.79\% & \textbf{0.3312} & \textbf{39.06\%} & 0.3795 & 50.18\% \\
E17 & 78D $[X, V]$ & Exploratory Best-per-Task Specialist Architectures & 1,002,831 & \textbf{0.6453} & \textbf{65.37\%} & 0.3141 & 37.76\% & 0.4738 & 53.41\% \\
\bottomrule
\end{tabular}%
}
\vspace{0.3em}
\noindent\footnotesize\textit{Note:} Experiment identifiers follow the chronological experiment registry. E8 was an intermediate exploratory configuration combining 117D $[X,V,A]$ with learned attention pooling under standard cross-entropy. We omit it from the primary table because E7 and E9 outperformed it (Stroke Macro-$F_1 = 0.6277$, Direction Macro-$F_1 = 0.2634$) and it did not reduce acceleration derivative noise. E13 was reserved in planning for a hybrid Transformer-BiGRU architecture and was never run, because the recurrent baselines (E10, E11) showed no advantage over the Transformer trunk. Complete specifications for E0--E17 are in the project experiment registry.
\end{table*}

\begin{figure}[t]
\centering
\includegraphics[width=\columnwidth]{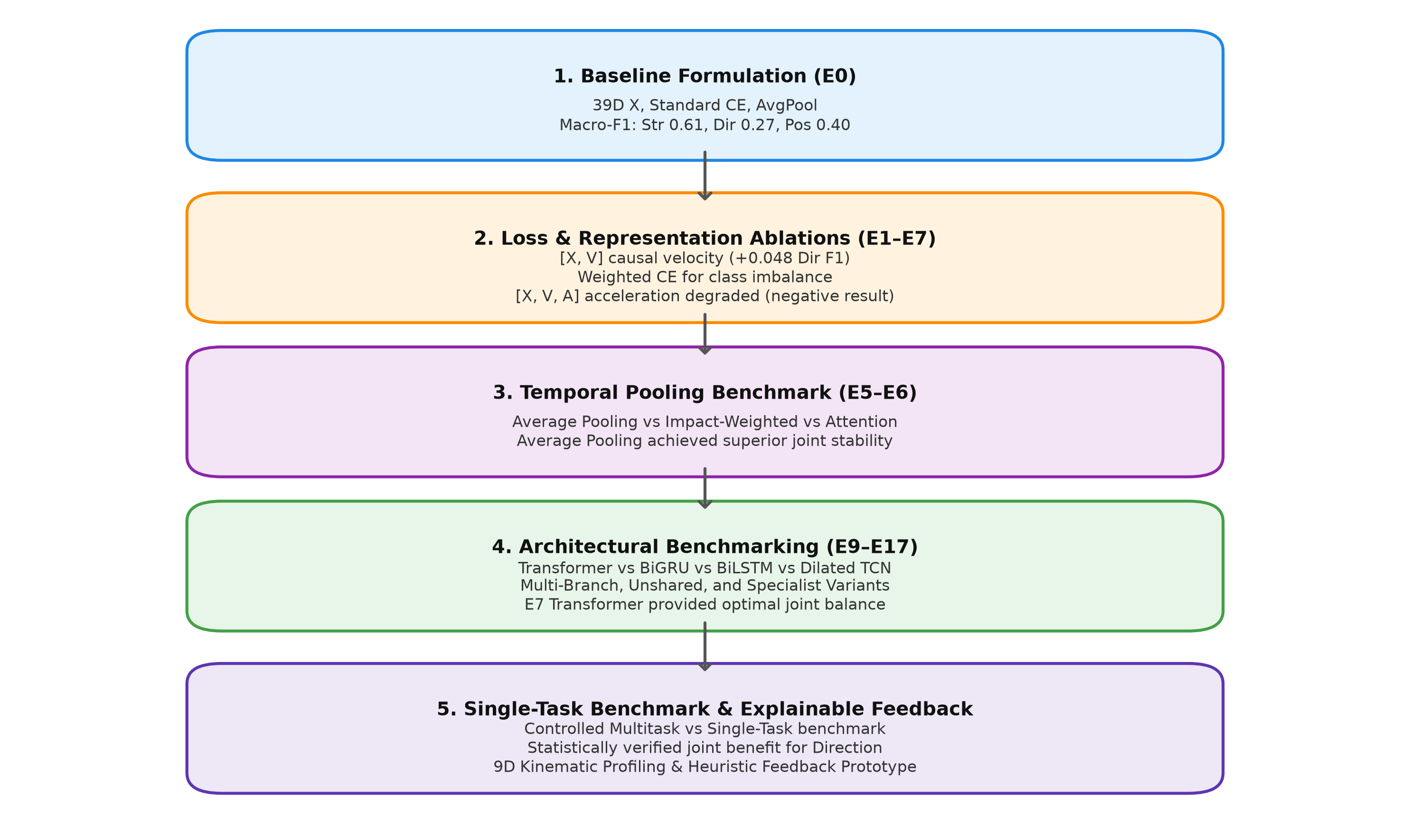}
\caption{Systematic experimental research journey and controlled decision points across representations, losses, pooling, and temporal architectures.}
\label{fig:research_journey}
\end{figure}

The ablation sequence gives these main findings:
\begin{itemize}
    \item \textbf{Causal Velocity (E0 vs. E3):} Adding first-difference velocities to the spatial keypoints ($[X, V]$) raised Direction Macro-$F_1$ from $0.2665$ to $0.3012$ and Stroke Type Macro-$F_1$ from $0.6096$ to $0.6380$. Motion information appears to help stroke characterization.
    \item \textbf{Acceleration Hurts (E4 Negative Result):} Adding second-difference acceleration features ($[X, V, A]$, $117$ dimensions) cut performance substantially (Stroke Macro-$F_1$ fell to $0.5821$; Direction to $0.2745$). In monocular video, numerical differentiation amplifies high-frequency tracking jitter, so second derivatives are mostly noise.
    \item \textbf{Class Weighting (E0 vs. E1):} Inverse-frequency class weighting eased minority-class starvation in Direction and lifted Macro-$F_1$ from $0.2665$ to $0.2982$, with no sign of added instability. Focal loss (E2) did less ($0.2814$).
    \item \textbf{Temporal Pooling (E5, E6, E9 vs. E7):} Neither Gaussian impact weighting (E5, Direction Macro-$F_1 = 0.2627$) nor learned attention pooling (E6 on 39D: $0.2649$; E9 on 78D: $0.2898$) beat simple Adaptive Average Pooling (E7: $0.3141$). Average pooling may help because it passes gradient evenly to all 30 frames.
    \item \textbf{Cumulative E7 Progression:} Direction Macro-$F_1$ rose from $0.2665$ in E0 to $0.3141$ in E7, a $17.9\%$ relative increase within the E-series exploratory comparison.
\end{itemize}

\subsection{Temporal Architecture Benchmark (E9--E17)}
The sequence architectures trade off differently across tasks:
\begin{itemize}
    \item The recurrent baselines (BiGRU E10 and BiLSTM E11) scored below the Transformer on Stroke Type ($0.5577$ and $0.5944$ vs. $0.6453$).
    \item The Dilated Residual TCN (E12) had the best Posture Quality result (Macro-$F_1 = 0.4738$, BalAcc = $53.41\%$), but did poorly on Stroke Type ($0.5354$) and Direction ($0.2676$).
    \item The multi-branch (E14) and unshared (E15) configurations nearly tripled the parameter count ($>1.5\text{M}$ parameters) without a matching gain in joint performance.
\end{itemize}
We chose E7 as the downstream configuration because it balanced the three tasks best within a compact 569k-parameter footprint.

\subsection{Canonical Multi-Task Downstream Results (E7)}
\begin{figure*}[t]
\centering
\includegraphics[width=\textwidth]{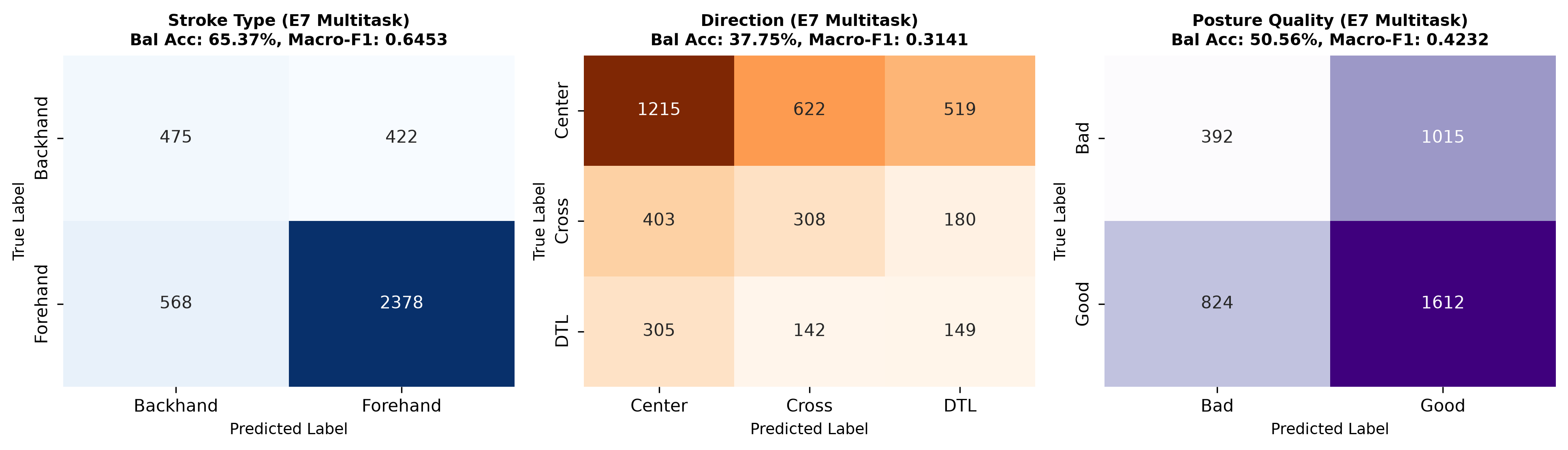}
\caption{Pooled confusion matrices for E7 Multitask across all 21 outer folds under Clean LOPO v2 ($N=3,843$ stroke evaluations). Confusion matrices show substantial class-wise variation under weighted cross-entropy.}
\label{fig:downstream_cms}
\end{figure*}

Table~\ref{tab:per_class_results} and Figure~\ref{fig:downstream_cms} report per-class metrics and pooled confusion matrices for the canonical E7 model across all 21 outer folds ($N = 3,843$ total test evaluations).

\begin{table}[h]
\centering
\small
\caption{Canonical per-class performance metrics for E7 Multitask under Clean LOPO v2 (pooled across 21 outer evaluations, $N=3,843$). Table~4 pools predictions across the 21 outer evaluations before computing per-class metrics, whereas Tables~3 and~5 summarize player-level metrics with the three seeds averaged within each player. Because these are different aggregation procedures, Macro-$F_1$ and balanced-accuracy values computed from Table~4 are not expected to equal the player-averaged values reported in Tables~3 and~5.}
\label{tab:per_class_results}
\resizebox{\columnwidth}{!}{%
\begin{tabular}{llccc}
\toprule
\textbf{Task} & \textbf{Class} & \textbf{Precision} & \textbf{Recall} & \textbf{$F_1$-Score} \\
\midrule
\multirow{2}{*}{\textbf{Stroke Type}} 
& Backhand & 0.4554 & 0.5295 & 0.4897 \\
& Forehand & 0.8493 & 0.8072 & 0.8277 \\
\midrule
\multirow{3}{*}{\textbf{Direction}} 
& Center        & 0.6322 & 0.5157 & 0.5680 \\
& Cross-Court   & 0.3169 & 0.3457 & 0.3307 \\
& Down-the-Line & 0.1757 & 0.2500 & 0.2064 \\
\midrule
\multirow{2}{*}{\textbf{Posture Quality}} 
& Bad  & 0.3224 & 0.2786 & 0.2989 \\
& Good & 0.6136 & 0.6615 & 0.6366 \\
\bottomrule
\end{tabular}%
}
\end{table}

\subsection{Controlled Single-Task vs. Multi-Task Benchmark}
\begin{figure}[t]
\centering
\includegraphics[width=\columnwidth]{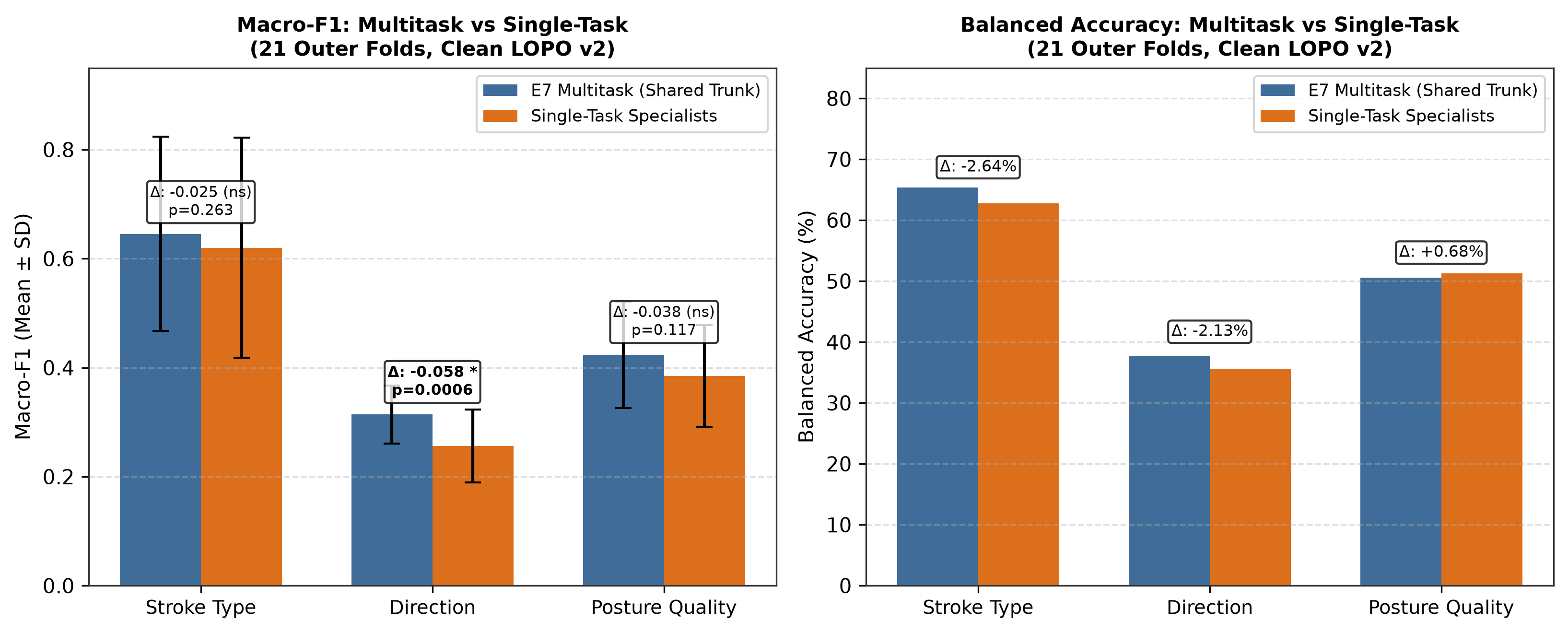}
\caption{Comparison of Macro-$F_1$ and Balanced Accuracy between E7 Multitask and Single-Task Specialists across 21 outer folds under Clean LOPO v2.}
\label{fig:single_vs_multi}
\end{figure}

Table~\ref{tab:single_vs_multi_results} and Figure~\ref{fig:single_vs_multi} give the paired comparison between the canonical multi-task E7 model and the independently trained single-task specialists (S1, S2, S3) under Clean LOPO v2.

\begin{table*}[t]
\centering
\small
\caption{Controlled single-task vs. multi-task benchmark under Clean LOPO v2 ($N=7$ independent players, averaged across 3 seeds). Differences ($\Delta = \text{Single} - \text{Multitask}$) are evaluated using exact paired sign-flip permutation tests ($2^7 = 128$ permutations) and player-level bootstrap confidence intervals. The Direction difference provides suggestive evidence ($p = 0.0469$, player-level 95\% CI: $[-0.0967, -0.0194]$) based on seven independent players, but does not meet a conservative Bonferroni-corrected threshold ($0.05/3 \approx 0.0167$).}
\label{tab:single_vs_multi_results}
\resizebox{\textwidth}{!}{%
\begin{tabular}{llcccccc}
\toprule
\textbf{Task} & \textbf{Model Paradigm} & \textbf{Accuracy} & \textbf{BalAcc} & \textbf{Macro-$F_1$} & \textbf{Mean $\Delta$ (F1)} & \textbf{95\% Player Bootstrap CI} & \textbf{Paired Permutation $p$} \\
\midrule
\multirow{2}{*}{\textbf{Stroke Type}} 
& E7 Multitask (Shared Trunk) & \textbf{74.27\%} & \textbf{65.37\%} & \textbf{0.6453} & \multirow{2}{*}{$-0.0253$} & \multirow{2}{*}{$[-0.0785, +0.0245]$} & \multirow{2}{*}{$p = 0.4062$ (ns)} \\
& S1 Single-Task Specialist   & 70.76\% & 62.73\% & 0.6200 & & & \\
\midrule
\multirow{2}{*}{\textbf{Direction}} 
& E7 Multitask (Shared Trunk) & \textbf{43.90\%} & \textbf{37.75\%} & \textbf{0.3141} & \multirow{2}{*}{\textbf{$-0.0578$}} & \multirow{2}{*}{\textbf{$[-0.0967, -0.0194]$}} & \multirow{2}{*}{\textbf{$p = 0.0469$ (suggestive)}} \\
& S2 Single-Task Specialist   & 30.02\% & 35.62\% & 0.2563 & & & \\
\midrule
\multirow{2}{*}{\textbf{Posture Quality}} 
& E7 Multitask (Shared Trunk) & 52.13\% & 50.56\% & \textbf{0.4232} & \multirow{2}{*}{$-0.0382$} & \multirow{2}{*}{$[-0.0764, +0.0059]$} & \multirow{2}{*}{$p = 0.1406$ (ns)} \\
& S3 Single-Task Specialist   & \textbf{56.03\%} & \textbf{51.24\%} & 0.3850 & & & \\
\bottomrule
\end{tabular}%
}
\end{table*}

Task by task:
\begin{enumerate}
    \item \textbf{Stroke Type:} The single-task specialist S1 reaches Macro-$F_1 = 0.6200$, against $0.6453$ for E7 ($\Delta = -0.0253$). The 95\% player-level confidence interval spans zero ($[-0.0785, +0.0245]$, exact permutation $p = 0.4062$), so we find no significant difference between dedicated and joint training.
    \item \textbf{Direction:} The single-task specialist S2 does worse, with Macro-$F_1 = 0.2563$ against $0.3141$ for E7 ($\Delta = -0.0578$). Under the paired exact sign-flip permutation test across the 7 players, this gap is suggestive evidence of a multitask effect ($p = 0.0469$, player-level 95\% CI: $[-0.0967, -0.0194]$). It rests on only seven players and does not survive Bonferroni correction for three task comparisons ($0.05/3 \approx 0.0167$), so it should be read with caution. The Direction specialist was also unstable at the class level under heavy imbalance: its recall on the majority Center class was only $18.39\%$, although Center makes up about $61.3\%$ of strokes. That may reflect shifted decision boundaries under class-weighted optimization, but this experiment does not isolate the mechanism.
    \item \textbf{Posture Quality:} The single-task specialist S3 reaches Macro-$F_1 = 0.3850$ versus $0.4232$ for E7 ($\Delta = -0.0382$). The difference is not significant ($p = 0.1406$, 95\% CI: $[-0.0764, +0.0059]$).
\end{enumerate}

\subsection{Explainable Kinematic Profiling and Feedback Demonstration}
\begin{figure}[t]
\centering
\includegraphics[width=\columnwidth]{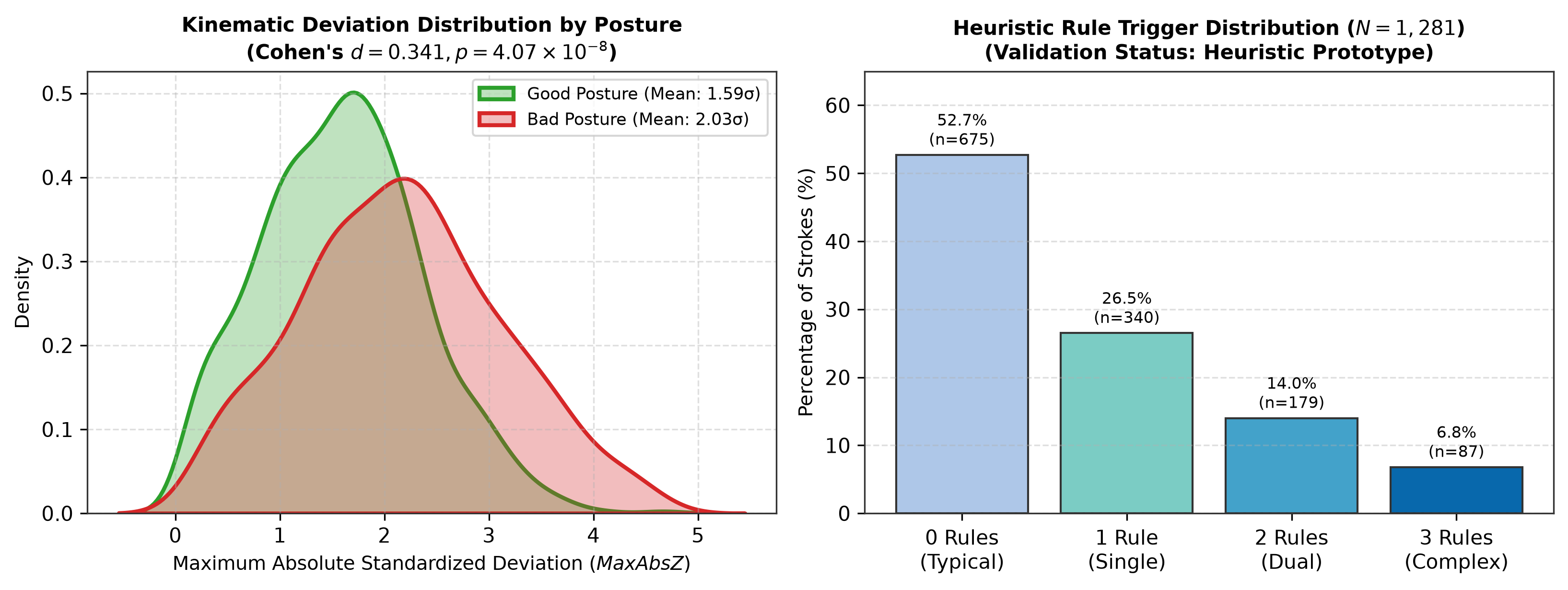}
\caption{Kinematic deviation analysis: (Left) Distribution of maximum absolute z-scores ($MaxAbsZ$) comparing good vs. bad posture strokes. (Right) Heuristic coaching rule trigger distribution across $1,281$ strokes.}
\label{fig:kinematic_pipeline}
\end{figure}

Across all $1,281$ labeled strokes, the 9-dimensional kinematic deviation profiles separate good from bad posture strokes to a measurable but modest degree (Figure~\ref{fig:kinematic_pipeline}):
\begin{itemize}
    \item \textbf{Overall Kinematic Deviation ($MaxAbsZ$):} Bad posture strokes had higher maximum standardized deviations ($2.03\sigma \pm 0.98$) than good posture strokes ($1.59\sigma \pm 0.76$), with absolute Cohen's $|d| = 0.341$ and nominal $p = 4.07 \times 10^{-8}$.
    \item \textbf{Mean Absolute Deviation ($MAD_z$):} Bad posture strokes had a larger average spread across joints ($0.80\sigma$ vs. $0.70\sigma, |d| = 0.328$, nominal $p = 9.42 \times 10^{-6}$).
    \item \textbf{Kinetic Sequencing Lag:} Bad-posture strokes had lower standardized sequencing-lag values than good-posture strokes ($-0.12\sigma$ vs. $+0.05\sigma, |d| = 0.167$, nominal $p = 0.00136$).
\end{itemize}
These p-values are nominal and were not adjusted for multiple comparisons, so we treat the results as exploratory and not confirmatory. Several engineered kinematic dimensions also reflect concepts in the posture annotation rubric (Section~\ref{sec:dataset}). The separation therefore shows that the engineered features agree with the existing annotation scheme, not that they independently identify technical proficiency.

The feedback layer has 30 deterministic heuristic rules and is treated as heuristic/unvalidated. The manual audit of $N=50$ cases checked logical behavior, not clinical or expert validity. Across the dataset, $52.7\%$ of strokes were kinematically typical ($0$ rules triggered), $26.5\%$ triggered $1$ rule, $14.0\%$ triggered $2$ rules, and $6.8\%$ triggered $3$ rules. Rules fired more often on strokes labeled as bad posture ($55.0\%$) than on good posture ($42.9\%$). This is preliminary evidence that the rules track the existing posture annotations, but it is not expert validation of the feedback content. Table~\ref{tab:representative_rules} lists representative rules that map standardized kinematic deviations to explanatory coaching feedback; the complete 30-rule decision table is in Supplementary Table S1.

\begin{table*}[t]
\centering
\small
\caption{Representative heuristic feedback rules across key kinematic categories. All rules carry the explicit status `heuristic\_unvalidated' pending prospective validation by qualified tennis coaches. The complete 30-rule decision table is detailed in Supplementary Table S1.}
\label{tab:representative_rules}
\resizebox{\textwidth}{!}{%
\begin{tabular}{lllllp{0.36\textwidth}c}
\toprule
\textbf{Rule ID} & \textbf{Kinematic Feature} & \textbf{Deviation} & \textbf{Context} & \textbf{Category} & \textbf{Generated Explanatory Feedback Statement} & \textbf{Validation Status} \\
\midrule
\texttt{RULE\_KNEE\_LOAD\_NEG\_STR} & Prep Knee Flexion ($^{\circ}$) & $z \le -2.0$ & Any & Lower-Body & Consider inspecting knee bend depth and lower-body engagement prior to forward swing initiation. & \texttt{heuristic\_unvalidated} \\
\texttt{RULE\_KNEE\_THRUST\_NEG\_MOD} & Knee Extension Thrust ($^{\circ}$) & $-2.0 < z \le -1.5$ & Any & Leg Drive & Consider reviewing vertical leg drive and upward push-off timing into ball contact. & \texttt{heuristic\_unvalidated} \\
\texttt{RULE\_XFACTOR\_PREP\_NEG\_STR} & Prep Max X-Factor ($^{\circ}$) & $z \le -2.0$ & Any & Torso Coil & Consider inspecting trunk coiling mechanics; evaluate whether backswing relies predominantly on arm draw. & \texttt{heuristic\_unvalidated} \\
\texttt{RULE\_ELBOW\_NEG\_STR} & Contact Elbow Angle ($^{\circ}$) & $z \le -2.0$ & Any & Contact Spacing & Consider inspecting ball spacing and strike-zone positioning to prevent jammed contact. & \texttt{heuristic\_unvalidated} \\
\texttt{RULE\_ARM\_EXT\_NEG\_MOD} & Arm Extension Ratio & $-2.0 < z \le -1.5$ & Any & Reach Radius & Consider reviewing lateral spacing from the body and extension through the strike zone. & \texttt{heuristic\_unvalidated} \\
\texttt{RULE\_SEQ\_LAG\_NEG\_STR} & Hip-Wrist Lag (frames) & $z \le -2.0$ & Any & Sequencing & Consider inspecting kinetic chain sequencing in video; verify that hips lead forward swing initiation. & \texttt{heuristic\_unvalidated} \\
\bottomrule
\end{tabular}%
}
\end{table*}

\begin{figure*}[t]
\centering
\includegraphics[width=\textwidth]{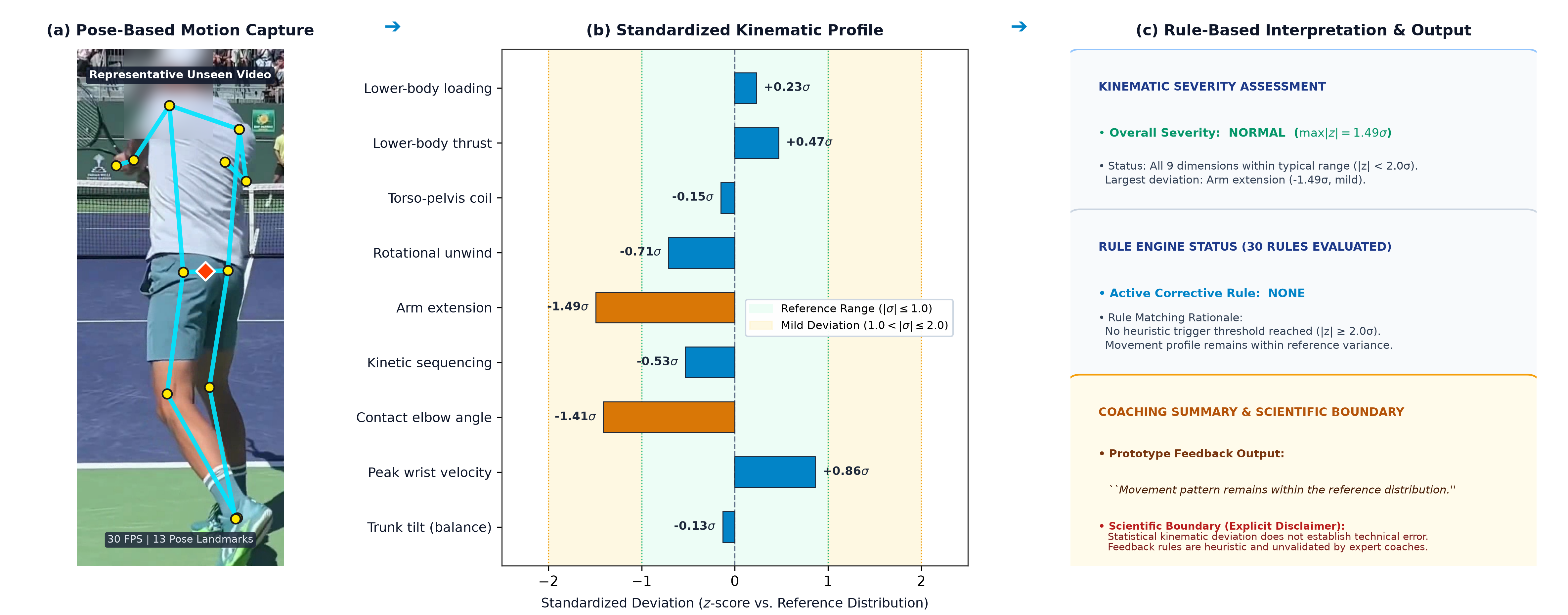}
\caption{Representative end-to-end example of the explainable kinematic profiling pipeline on an unseen external video. (a) A representative stroke frame with pose landmarks. (b) Standardized deviations across selected kinematic dimensions relative to the player-disjoint reference distribution. (c) Rule-based interpretation and feedback generation. The example illustrates how kinematic deviations are converted into cautious human-readable feedback; the rule-based feedback layer is heuristic and has not been validated by expert coaches.}
\label{fig:feedback_demonstration}
\end{figure*}

Figure~\ref{fig:feedback_demonstration} shows a representative example on an unseen video. The system extracts pose landmarks, derives the nine kinematic dimensions, and standardizes them against the player-disjoint reference distribution. The deterministic rule engine then processes the resulting deviations. Here, arm extension has the largest negative deviation ($-1.49\sigma$), and the other displayed dimensions sit closer to the reference distribution. No rule threshold for corrective feedback is met, so the system gives no active correction. The example shows how the pipeline is meant to be read; it does not validate the generated coaching interpretation.
\section{Discussion}
\label{sec:discussion}

\subsection{Architectural Synthesis and Negative Results}
Greater architectural complexity did not reliably improve cross-player generalization. In sports computer vision it is tempting to add higher-order kinetic derivatives, dense attention mechanisms, or large multi-branch networks. Our controlled ablations produced several negative results that are useful as engineering guidance:

\begin{enumerate}
    \item \textbf{Noise in Second-Order Derivatives:} Adding first-difference causal velocities ($[X, V]$) together with class weighting (E7) raised Direction Macro-$F_1$ by $+17.9\%$ relative to E0. Adding second-difference acceleration features ($[X, V, A]$, E4) degraded all three tasks substantially. A likely explanation is that landmark jitter in monocular video grows under repeated numerical differentiation and injects high-frequency noise that disrupts temporal attention weights, though we did not test this directly.
    \item \textbf{Average Pooling over Attention:} We expected learned attention pooling (E6) and Gaussian impact weighting (E5) to focus better on the contact frame. Standard Adaptive Average Pooling (E7) generalized better across players. The exact impact instant varies across players and camera angles, and average pooling may do better because it keeps gradient flow to every frame and captures the broader kinematic arc.
    \item \textbf{Task-Specific Trade-offs in Convolutional Models:} The Dilated Residual TCN (E12) did well on Posture Quality (Macro-$F_1 = 0.4738$), ahead of the Transformer backbone ($0.4232$), but was much worse on Stroke Type ($0.5354$ vs. $0.6453$). Local convolutional receptive fields may suit localized joint configuration errors, while the global temporal interactions in self-attention may matter more for stroke type on this dataset.
    \item \textbf{Parameter Redundancy in Unshared Encoders:} The fully unshared architecture (E15, $1.52\text{M}$ parameters) and the multi-branch network (E14, $1.50\text{M}$ parameters) did not improve on the compact shared encoder (E7, $569\text{k}$ parameters). This is consistent with the three tasks benefiting from partly shared anatomical representations.
\end{enumerate}

\subsection{Why Multi-Task Learning Benefits Direction}
We tested whether joint training beats independent task-specific training in a controlled single-task benchmark across all 21 outer folds under Clean LOPO v2. Across the 7 independent players, we found no statistically significant difference between joint and independently trained models for Stroke Type ($p = 0.4062$) or Posture Quality ($p = 0.1406$). For Direction, independent training lowered Macro-$F_1$ ($\Delta = -0.0578$, exact paired permutation $p = 0.0469$, player-level 95\% CI: $[-0.0967, -0.0194]$). That is suggestive evidence of a multitask regularization effect at the player level. With only seven players it does not clear a conservative Bonferroni threshold ($0.05/3 \approx 0.0167$), so it should be read with caution.

A regularization effect under heavy class imbalance would fit this pattern. On this dataset, Direction is hard to predict from body kinematics alone: telling cross-court from down-the-line shots depends on subtle differences in contact point and wrist angle, which individual player style easily swamps. Direction Balanced Accuracy is $37.75\%$ against a 3-class chance level of $33.33\%$, and under inverse-frequency weighting the single-task model had unstable class boundaries, with Center recall falling to $18.39\%$ despite $61.3\%$ prevalence. In the multi-task model, the shared Transformer backbone also receives supervision from Stroke Type and Posture Quality. This is consistent with shared representation learning acting as a useful regularizer for Direction under heavy imbalance, though our data do not show that it is the cause.

\subsection{Qualitative Comparison with Prior Literature}
Table~\ref{tab:literature_comparison} compares our framework with the study of Shimizu et al.~\cite{shimizu2019prediction} on pipeline design, task scope, and evaluation protocol.

\begin{table*}[t]
\centering
\small
\caption{Qualitative comparison of pipeline design, task scope, and evaluation protocols with prior pose-based tennis analysis.}
\label{tab:literature_comparison}
\resizebox{\textwidth}{!}{%
\begin{tabular}{p{0.22\textwidth} p{0.36\textwidth} p{0.38\textwidth}}
\toprule
\textbf{Analytical Dimension} & \textbf{Shimizu et al.~\cite{shimizu2019prediction}} & \textbf{This Work (TennisTransformer E7 Framework)} \\
\midrule
Pose Extraction Framework & OpenPose (Multi-stage 2D CNN) & MediaPipe Pose Landmarker (Top-down 3D model) \\
Coordinate Geometry & 2D image pixel coordinates & 3D hip-relative spatial coordinates ($\mathbf{X} \in \mathbb{R}^{39}$) \\
Temporal Representation & Handcrafted kinematic summary & Causal velocity augmented sequence ($[X, V] \in \mathbb{R}^{30 \times 78}$) \\
Stroke Segmentation & Manual temporal pre-annotation & Fully automated kinetic event detector ($F_1 = 0.8565$) \\
Temporal Modeling & LSTM on 2D pose and court position & 4-layer Transformer Encoder with self-attention \\
Task Coverage & Direction Prediction (1 primary task) & Stroke Type, Direction, Posture Quality, Kinematic Profiling (4 analytical outputs) \\
Evaluation Split Paradigm & Video-level split & Strict Clean LOPO v2 (Leave-One-Player-Out cross-validation) \\
Evaluation Metrics & Standard classification accuracy & Balanced Accuracy, Macro-$F_1$, and per-class metrics \\
Reported Direction Accuracy & $\sim 85\text{--}90\%$ (Standard accuracy on broadcast) & $43.90\%$ Accuracy, $37.75\%$ Balanced Acc, $0.3141$ Macro-$F_1$ \\
Hardware Feasibility & High-end GPU workstation & Consumer edge / CPU execution (Apple Silicon multi-processing) \\
\bottomrule
\end{tabular}%
}
\end{table*}

The numbers are not directly comparable. Shimizu et al. evaluated on high-resolution broadcast television footage using standard accuracy, whereas our results come from uncurated internet video under strict Leave-One-Player-Out cross-validation. Under player-disjoint splits, our framework adds automated event segmentation and covers more tasks, without requiring manual pre-trimming.

\subsection{Statistical Rarity versus Biomechanical Error}
Our kinematic profiling shows that \textit{statistical rarity is not the same as technical error}. Some elite strokes had extreme $z$-score deviations; for example, one player in the cohort had peak wrist velocities more than $8.50\sigma$ above the reference means. Such outliers most likely reflect exceptional kinetic acceleration, not faulty technique. Automated anomaly detection should therefore not be treated as a diagnostic engine without explicit expert adjudication.
\section{Limitations and Future Work}
\label{sec:limitations}

This section covers what the study does not establish.

\subsection{Coordinate Calibration and Monocular Depth Constraints}
MediaPipe Pose outputs 3D coordinates inferred by the model and scaled to an estimated torso bounding box. Without calibrated multi-camera rigs, known court fiducials, or physical anthropometric measurements, these coordinates \textit{lack SI unit calibration} (e.g., metric millimeters or radians verified by optical markers). Depth along the optical camera axis is also less certain than lateral or vertical displacement. The video corpus is mostly baseline and elevated broadcast views with some lateral-quarter angles. We did not stratify by viewpoint or model camera orientation explicitly. Kinematic profiles should therefore be read as view-normalized geometric approximations, not medical-grade gait or motion capture data.

\subsection{Annotation Subjectivity and Discrete Categorization}
One annotator labeled posture quality using qualitative biomechanical criteria. There was no multi-expert panel adjudication and no inter-rater agreement metric (e.g., Cohen's $\kappa$), so label reliability is unknown. The binary Good vs. Bad scheme also flattens a continuous range of athletic execution. Future work should use multi-expert panels with continuous visual analog scales or rubric-based scoring.

\subsection{Direction Prediction from Kinematics Alone}
Shot direction remains difficult under player-disjoint evaluation (Balanced Accuracy $37.75\%$, Macro-$F_1 = 0.3141$). Our framework infers direction only from human kinematic trajectories, whereas the ball's path in actual play depends on racket face angle, string tension, impact velocity, and incoming ball spin during a very brief contact window. A 30-frame full-body pose sequence with no ball tracking, court boundary localization, or racket head tracking cannot be expected to fully resolve shot direction. Pairing pose sequences with ball trajectory tracking (e.g., TrackNet) and court calibration is the obvious next step.

\subsection{Absence of Tactical Context and Optimal Shot Ground Truth}
We make no claim to select optimal shots or recommend tactics. Shot choice depends on opponent court position, ball bounce height, rally tempo, score state, and player fatigue, and our dataset has no tactical ground truth, opponent tracking, or rally progression data. The heuristic feedback prototype evaluates movement patterns only, not tactical decisions.

\subsection{Unvalidated Heuristic Feedback Layer}
The 30-rule feedback prototype shows that structured coaching observations can be generated from kinematic deviation profiles. It has not been prospectively validated with certified tennis coaches or sports medicine practitioners, and should be read strictly as a heuristic proof-of-concept.

\subsection{Future Research Directions}
Planned work includes:
\begin{enumerate}
    \item Combining high-speed contact-window modeling (2–3 frames around impact) with full-body sequences.
    \item Fusing player pose with multi-modal ball tracking and court boundary calibration.
    \item Scoring action quality on a continuous scale, with multi-rater expert adjudication.
    \item Testing cross-viewpoint generalization on calibrated multi-angle camera setups.
    \item Running formal prospective trials that compare automated feedback with certified coach interventions.
\end{enumerate}
\section{Conclusion}
\label{sec:conclusion}

This paper presented a player-disjoint framework for automated tennis stroke analysis from monocular RGB video. It combines a kinetic stroke event detector, a 78-dimensional causal velocity representation ($[X, V]$), a compact 569k-parameter multi-task Transformer (E7), exploratory kinematic profiling, and a heuristic feedback engine. No manual pre-annotation is needed, and evaluation avoids random splits that leak player identity.

On 206 frozen consensus impact events, adjudicated independently, the Composite Detector reached an $F_1$-score of $0.8565$ ($+3.24\%$ over baseline thresholding) with a mean absolute timing error of $80.6$\,ms, which is sufficient to segment untrimmed footage automatically. Under strict Clean LOPO v2 Leave-One-Player-Out cross-validation across 21 outer folds, the E7 Multi-Task Transformer gave the following player-disjoint results under substantial class imbalance: Stroke Type (Accuracy $74.27\%$, Balanced Accuracy $65.37\%$, Macro-$F_1 = 0.6453$), Direction (Accuracy $43.90\%$, Balanced Accuracy $37.75\%$, Macro-$F_1 = 0.3141$), and Posture Quality (Accuracy $52.13\%$, Balanced Accuracy $50.56\%$, Macro-$F_1 = 0.4232$).

Against independently trained task-specific models, multi-task training made no statistically significant difference for Stroke Type ($p = 0.4062$) or Posture Quality ($p = 0.1406$). Direction was the exception, with suggestive evidence of an advantage ($\Delta = -0.0578$, exact paired sign-flip permutation $p = 0.0469$, player-level 95\% CI: $[-0.0967, -0.0194]$). Since this rests on only seven independent players, we treat it cautiously. A possible explanation is that the shared representation regularizes Direction prediction under severe class imbalance, but that remains a hypothesis. The exploratory 9-dimensional kinematic deviation profiling also showed nominal, unadjusted separation between good and bad posture strokes ($|d| = 0.341$, nominal $p = 4.07 \times 10^{-8}$). We used this to generate heuristic coaching cards, which have no clinical or expert biomechanical validation.

The model does generalize to unseen players under strict Leave-One-Player-Out evaluation, with limits. Monocular pose estimation, direction inference from body motion alone, single-annotator posture labels, and unvalidated heuristic feedback all constrain what the results show. The framework is a research prototype for video-based athletic assistance. It is not a substitute for qualified coaching or laboratory-grade biomechanical measurement.
\section{Data and Code Availability}
\label{sec:data_code_availability}

Source code, experiment configurations, evaluation scripts, and the supplementary rule-engine specification will be released in a public repository upon publication. We cannot share the original videos, because they show identifiable individuals and yield video-derived human pose sequences. Anonymized derived features and experiment metadata may be released if the data-use and consent terms permit. Supplementary Table S1 gives the full 30-rule heuristic feedback decision engine.

\bibliographystyle{IEEEtran}
\bibliography{references}

\end{document}